\documentclass{Interspeech2024}

\interspeechcameraready

\usepackage{algorithm}
\usepackage[noend]{algpseudocode}
\usepackage{multirow}

\usepackage[T1]{fontenc}
\usepackage[utf8]{inputenc}

\title{Multi-stage Large Language Model Correction for Speech Recognition}

\name{Jie Pu, Thai-Son Nguyen, and Sebastian St{\"u}ker}

\address{Zoom Video Communications, Karlsruhe, Germany}


\begin{document}

\maketitle

\begin{abstract}
    
    In this paper, we investigate the usage of large language models (LLMs) to improve the performance of competitive speech recognition systems. Different from previous LLM-based ASR error correction methods, we propose a novel multi-stage approach that utilizes uncertainty estimation of ASR outputs and reasoning capability of LLMs. Specifically, the proposed approach has two stages: the first stage is about ASR uncertainty estimation and exploits N-best list hypotheses to identify less reliable transcriptions; The second stage works on these identified transcriptions and performs LLM-based corrections. This correction task is formulated as a multi-step rule-based LLM reasoning process, which uses explicitly written rules in prompts to decompose the task into concrete reasoning steps. Our experimental results demonstrate the effectiveness of the proposed method by showing 10\% $\sim$ 20\% relative improvement in WER over competitive ASR systems --- across multiple test domains and in zero-shot settings. 
\end{abstract}

\section{Introduction}

Large language models (LLMs) such as ChatGPT \cite{DBLP:journals/corr/abs-2303-08774}, Llama \cite{touvron2023llama} have changed the landscape of AI research because of their groundbreaking capabilities. In this paper, we focus on the topic of ASR error correction and explore the usage of LLMs to push the state-of-the-art (SOTA) performance. ASR correction has a long history in the community and serves as a post-processing to improve the readability and quality of ASR transcriptions. With the recent rise of LLMs, the LLM-based ASR correction methods have been proposed \cite{ma2023n, ma2023can, yang2023generative, radhakrishnan2023whispering, chen2023hyporadise}. These methods can be roughly categorized into two groups depending on whether to re-train LLMs: i) fine-tuning LLMs \cite{ma2023n, radhakrishnan2023whispering} and ii) in-context learning of LLMs \cite{min2022rethinking}, which utilizes prompts without changing parameters of LLMs \cite{ma2023can, yang2023generative}. Herein, we focus on the in-context learning scheme of LLMs, as it is much simpler (requiring no training) and avoid the over-fitting issue when fine-tuning \cite{radhakrishnan2023whispering, chen2023hyporadise}. 

Along with this line of research, different LLMs have been explored, e.g. T5 \cite{ma2023n}, Llama \cite{radhakrishnan2023whispering} and ChatGPT \cite{ma2023can}. Despite interesting results obtained in these works, the previous LLM-based approaches struggle to improve competitive ASR systems and cannot go beyond SOTA. On LibriSpeech, when the word-error-rate (WER) is lower than 2$\%$, there is very limited improvement that previous methods can contribute \cite{ma2023can, chen2023hyporadise}. This is an inherent limitation caused by two issues: i) \textit{over-correction}: LLMs make many unnecessary changes to the input transcription, as it steers the sentence more towards written language, instead of a verbatim transcript for the input speech. Because of the discrepancy between spoken language and written language, LLM-based correction can hinder the fidelity of ASR transcriptions \cite{chen2024its}; ii) \textit{multi-step reasoning} challenge for LLMs. The task of ASR error correction itself is quite complex and requires a high level of reasoning for LLMs, e.g. where to pick the operating point of its correction, when to perform no correction if the sentence is correct and which words to replace if an error is found. Simple prompting techniques cannot fully capture this reasoning rationale or accommodate the diversity within the task, and thus cannot yield satisfactory results over SOTA. 

To address the first issue of \textit{over-correction}, we propose a confidence-based ASR uncertainty estimation stage to first detect less reliable (uncertain) transcriptions and only perform LLM-based corrections on these detected sentences. Contrary to existing approaches \cite{gekhman2022red, qiu2021learning, li2021confidence}, we extract the confidence and uncertainty information from N-best list hypotheses and identify them by thresholding the obtained confidence scores. For transcriptions that are believed to be less reliable, we will prompt their N-best list to LLMs for a later correction. 

For the second issue of \textit{multi-step reasoning}, it is a well-known challenge for LLMs \cite{dziri2023faith, srivastava2023beyond}. The widely-used prompting techniques, chain-of-thought (CoT) prompting \cite{wei2022chain} and the prefix \textit{`let's think step by step'} \cite{kojima2022large}, are not best suited for our ASR correction task. Because the complexity of the task cannot be fully accommodated in several hand-crafted exemplars that CoT relies on \cite{suzgun2022challenging, zhou2022least} and it is also hard for LLMs to figure out the reasoning rationale entirely by themselves \cite{kojima2022large, khot2022decomposed}. To accommodate this complexity, \cite{yang2023generative} proposed a task-activating prompting (TAP), which needs four rounds of Q and A to properly set up the correction task. Herein, we want to use just one well-guided prompt and enable the zero-shot setting for LLMs. To this end, we propose a multi-step rule-based prompt that leverages explicitly written rules to facilitate the reasoning process of LLMs on this task. These rules guide LLMs to break down such a complex task into intermediate reasoning steps and also set up constrictions during the LLM generation process.

\begin{figure*}[!htb]
  \centering
  {\includegraphics[width=1\hsize]{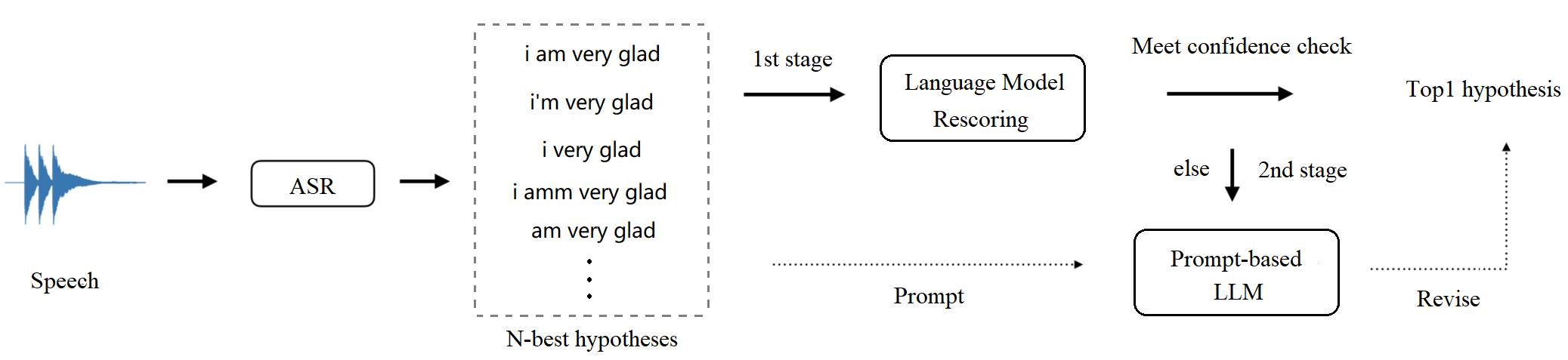}} 
  \caption{Overview of the multi-stage correction pipeline. The first stage is to detect less reliable transcriptions while the second stage performs a LLM-based correction. A confidence threshold will decide if transcriptions need to be sent to the second stage.} 
  \label{fig:pipeline}
\end{figure*}

Experiments showed with the proposed correction pipeline, we can achieve 10\% $\sim$ 20\% WER relative improvement consistently for competitive ASR systems. Our best system for the LibriSpeech benchmark can reach 1.3\% WER on LibriSpeech test-clean set, which sets a new state-of-the-art. Altogether, we summarize the main contribution and the novelty of this paper as following: 

\begin{itemize}
\item \textbf{ASR uncertainty estimation stage}. To deal with the over-correction issue of LLMs, we propose an uncertainty estimation to identify less reliable transcriptions. Given that, only a small portion of uncertain utterances will be sent to LLMs. It helps to keep the fidelity of majority spoken transcriptions and alleviates the total computational cost of using LLMs.

\item \textbf{Multi-step rule-based reasoning}. To facilitate the complex LLM reasoning process for ASR error correction, we propose a multi-step rule-based prompt that explicitly decomposes the task into concrete reasoning steps and constrictions to follow. With these explicitly defined rules, we sidestep the need to provide hand-crafted exemplars or craft fine-tuning datasets. Besides, it shows that LLMs can be zero-shot reasoners for the ASR correction task and  demonstrates the SOTA performance in zero-shot settings.

\end{itemize}

\section{Methodology}
\label{sec:method}
We propose a multi-stage ASR correction pipeline as Figure~\ref{fig:pipeline}. In the first stage we detect less reliable (uncertain) ASR transcriptions via thresholding obtained confidence scores. In the second stage, we perform LLM-based ASR corrections. On the whole, the objective is to send utterances with confidence scores below a threshold to LLMs, while keeping the majority of utterances on the first stage to gain quality, latency, and reliability.

\subsection{Stage 1: ASR Confidence and Uncertainty}
Traditionally, ASR confidence scores were widely used to evaluate the reliability of recognition results \cite{gekhman2022red}. Unlike hidden Markov model ASR systems, confidence scores of end-to-end ASR networks cannot be reliably obtained. For this, we propose to extract confidence and uncertainty information from N-best hypotheses. Compared to \cite{qiu2021learning, li2021confidence}, it is much simpler and requires no model training. Specifically, this stage contains two steps: i) LM re-scoring to provide scores for utterances by combining ASR models with a language model; ii) Softmax these scores and thresholding. 

\textbf{Language model re-scoring} 
\label{sec:re-ranking}
is a popular choice to estimate the quality of ASR transcriptions by exploiting an external language model \cite{chen2023large, huang2024multilingual}. Assume that we have an N-best list of hypotheses $\textbf{y}_{\textbf{1}}, \dots \textbf{y}_{\textbf{N}}$ which are generated from the decoding inference for an utterance $\textbf{x}$. In this stage, we employ a LM that provides a sentence-level probability for hypothesis $\textbf{y}_{\textbf{i}}$. Following the works in \cite{udagawa2022effect},
we calculate the score $Score (\textbf{y}_{\textbf{i}})$ used for re-scoring as :
\begin{align}
\label{equation_rerank}
&  Score (\textbf{y}_{\textbf{i}}) = \text{log} \hspace{0.5mm} \mathcal{P}_{\text{ASR}}(\textbf{y}_{\textbf{i}}|\textbf{x}) + \alpha \hspace{0.5mm} \text{log} \hspace{0.5mm} \mathcal{P}_{\text{LM}}(\textbf{y}_{\textbf{i}})
\end{align}
\noindent where $\mathcal{P}_{\text{LM}}(\textbf{y}_{\textbf{i}})$ is the LM probability for hypothesis $\textbf{y}_{\textbf{i}}$, and $\mathcal{P}_{\text{ASR}}(\textbf{y}_{\textbf{i}}|\textbf{x})$ is the probability obtained from the ASR model. The weight $\alpha$ are found using a development set.

\textbf{Softmax and thresholding}. Inspired by \cite{pearce2021understanding, hendrycks2016baseline}, we normalize the scores $Score (\textbf{y}_{\textbf{i}})$ of one N-best list with \textit{Softmax} function and simply evaluate $Score_{\text{best}}$, the highest normalized score. The softmax output, whose values are between 0 and 1, are interpreted as the confidence and  uncertainty measure. 
If larger than a threshold $\beta$, then this hypothesis is obtained with some confidence and more likely to contain no error. Otherwise, the hypothesis are uncertain and may require LLMs to correct.

The rationale behind is, we observed that when ASR transcriptions are correct, the highest $Score (\textbf{y}_{\textbf{i}})$ within an N-best list is often much larger than others, i.e. models are confident about this output and provide a large probability. After \textit{Softmax} function, its value tends to be bigger than the threshold $\beta$. On the other hand, when ASR transcriptions contain errors, it is often the case that models are uncertain about its output and produce several alternative words. The values of probabilities are then distributed across different candidates and result into no big values. After \textit{Softmax}, even the highest value is small.

\subsection{Stage 2: Large Language Model Correction}
\label{LLM correction}

In this stage, we perform a LLM-based correction. Different from previous methods on handling complex reasoning tasks \cite{yang2023generative, zhou2022least, khot2022decomposed}, we want to simplify the LLM prompting process and fully explore the zero-shot capability of LLMs given one single prompt, like \cite{kojima2022large}. Inspired by \cite{clark2021transformers} where rules can be stated in natural language and guide the LLM generation process, we formulate the problem of ASR correction as a multi-step rule-based LLM reasoning, where pre-defined rules decompose the problem into reasoning steps and set up constraints for the LLM generation. Besides, the proposed prompt exploits vast linguistic knowledge of LLMs and the word-level information across N-best lists to generate a final transcript. The detail of this prompt is in Algorithm~1, with rules highlighted as follows: 

\begin{algorithm}[!tb]
\small
\caption{Prompt for LLM}\label{Prompt}
\begin{algorithmic}[]
\vspace{1mm}
\State \textbf{Input}: \text{an N-best list from ASR, containing N hypotheses. } 
\text{They are ranked by scores during beam search} \textbf{y}[1] ... \textbf{y}[n].
\State \textbf{Prompt}: You are an excellent assistant for speech recognition system. Your task is to check and correct potential errors in speech transcriptions. \\\vspace{0.5mm}
Please follow the following rules, and here is the sentence to work on: \textbf{y}[1].\\\vspace{0.5mm}
You need to first consider the following variant sentences and try to pick corrected words from them: \textbf{y}[2] ... \textbf{y}[n]. \\\vspace{1mm}
Additional rules for this modification: \\ \vspace{1mm}
1. If any word in the original sentence looks weird or inconsistent, then replace it with a corresponding word from variant sentences.\\ \vspace{0.5mm}
2. You don’t have to modify the original sentence if it already looks good. \\ \vspace{0.5mm}
3. Keep the sentence structure and word order intact.\\ \vspace{0.5mm}
4. Only replace words in the original sentence with ones from variant sentences. Do not simply add or delete words.\\ \vspace{0.5mm}
5. Try to make the corrected sentence have the same number of words as the original sentence. \\ \vspace{0.5mm}
6. Ignore punctuation. \\ \vspace{0.5mm}
7. Use U.S. English. \\ \vspace{0.5mm}
8. Output only one modified sentence and no explanation. \\ \vspace{1mm}
\textbf{Output:} the revised hypothesis \textbf{\^{y}}[1]
\end{algorithmic}
\end{algorithm}
\begin{itemize}
\item \textbf{New words}. Rule 4 restricts LLMs to only use words from the N-best list, otherwise LLMs may use synonyms in the correction process. We notice that  LLMs such as ChatGPT tend to format ASR transcripts by adding conjunctions or removing repetitions, which results in more coherent sentences, however hinders the fidelity of speech transcriptions.

\item \textbf{Creativity}. Rules 3 and 5 are designed to confine the creativity and randomness of LLMs, 
by restricting the structure and length of its output sentence, so that the output will stay close to a verbatim transcript for the input speech.

\item \textbf{Output standardization}. Rule 7 can be changed to other English variants and spelling systems, e.g. U.K. English. Rule 6 is included for the convenience of ASR evaluation. This may be removed if downstream NLP tasks exist and can benefit from punctuation, e.g. translation or summarization.

\item \textbf{Explanation}. Rule 8 instructs LLMs to provide no explanation and keep their reasoning implicit. This is convenient for the output format and also appears in other works \cite{ma2023can}, e.g. \textit{you need to provide the corrected ASR hypothesis directly without any explanations}. However, this requirement introduces more challenges for LLMs to perform this task, because it prohibits LLMs to produce more tokens to \textit{think} while the intermedia reasoning process of LLMs is crucial when handling complex tasks \cite{dziri2023faith, mukherjee2023orca}. 
\end{itemize}

Compared with Zero-shot-CoT \cite{kojima2022large} that needs to prompt twice to first perform reasoning and then extract answer, our rule-based prompt enables LLMs to have their complex reasoning process implicit and output results directly. It is more efficient and demonstrates a higher level of reasoning capability empowered by our prompt.

\section{Experimental Setup}
\label{sec:Experiments}

\subsection{Data Sets}
\label{Data}
We used publicly available English datasets for experiments:
\begin{itemize}
\item \textbf{LibriSpeech} (\textbf{LS}) \cite{panayotov2015librispeech} is a collection of around 960 hours of read speech from audio books. We used the standard split for train, validation and test sets (test-clean, test-other).

\item \textbf{Common Voice} (\textbf{CV}) \cite{ardila2019common} consists of about 900 hours of English transcribed audio where speakers record text from Wikipedia. This data set has a large variation in quality and speakers, as anyone can submit recorded contributions.

\item \textbf{TED-LIUM 3} (\textbf{TL}) \cite{hernandez2018ted} contains 452 hours of speech from TED talks. This dataset represents presentation speech which is a popular domain nowadays.

\item \textbf{Multilingual LibriSpeech} (\textbf{MLS}) \cite{pratap2020mls} is an extension of LibriSpeech and contains 44.5K hours of English speech. It is for the benchmark when a large of mount of data available. 
\end{itemize}
\subsection{ASR Models} 
We extract 40 log-mel filterbank coefficients with mean normalization as the input features, and use SpecAugment~\cite{park2019specaugment} for data augmentation during training. Labels are generated from a sub-word tokenzier with the vocab size of 4000 units. The attention-based sequence-to-sequence ASR models were built and trained following \cite{nguyen2020improving}. In all experiments, we used the same encoder network consisting of two convolutional layers and six layers of bidirectional LSTMs with 1,280 cells, and the decoder network with two unidirectional LSTM layers with 1,280 cells. 

\subsection{Language Models} 
\label{language models}
For \textit{Stage 1}, we choose to use the GPT-J model \cite{wang2021gpt} across different test domains, instead of applying several in-domain language models. 
As stated in \cite{chen2023large, huang2024multilingual}, re-scoring with a generalized language model can achieve comparable or better performance than using one in-domain language model.

For \textit{Stage 2} we explored GPT-3.5 and GPT-4 with versions released on March 2023. Their optimal configurations have been investigated on two specific factors: 1) whether to allow new words outside of the N-best list, i.e. Rule 4 in the proposed prompt, and 2) the hyper-parameter \textit{temperature} in ChatGPT. Table~\ref{tab_chatgpt} shows a comparison between different configurations. We can see that GPT-4 performs better than GPT-3.5 due to its capability to handle more complex instructions in prompts \cite{DBLP:journals/corr/abs-2303-08774}. Lowering the value of \textit{temperature} reduces the randomness of GPT-4's output, thus helps to reduce the WER. The best performance is obtained when disallowing new words.
\begin{table}[!tp]
\centering
\caption{WERs in \% of different ChatGPT configurations on LibriSpeech dev-clean set. GPT-J and ChatGPT language models and the ASR model trained on LS and MLS, are used. 
}
\label{tab_chatgpt}
\resizebox{0.9\hsize}{!}{%
\begin{tabular}{|l|c|c|c|}
\hline
Model  & Temperature  & Allow new words                       & WER \\ \hline
GPT-3.5   &    0.7  & Y    & 2.6     \\ 
GPT-4     &    0.7  & Y    & 1.7    \\
GPT-4     &    0.5  & N    & 1.5     \\
GPT-4     &    0.2  & N    & \textbf{1.4} \\ \hline
\end{tabular}
}
\end{table}
\begin{table}[!tp]
\centering
\caption{WERs in \% of different N-best list sizes, the weight $\alpha$ and the confidence level threshold $\beta$ on LibriSpeech dev-clean set. GPT-J and ChatGPT language models and the ASR model trained on the multi-domain data (LS, CV and TL) are used. For each N, the optimal values of $\alpha$ and $\beta$ are presented here. Given the confidence threshold, a percentage \%  of total speech utterances will be sent to the stage 2. }
\label{tab_config}
\resizebox{1\hsize}{!}{%
\begin{tabular}{|c|c|c|c|c|}
\hline
N-best  & $\alpha$ &  $\beta$  & WER & $\%$ of utterances to Stage 2 \\ \hline
5   & 3.0 & 0.70    & \textbf{2.1}  &  23.0 \\ 
8   & 4.5 & 0.45    & 2.2   &  7.8 \\
16  & 4.5  & 0.60    & 2.2   &  19.7\\  \hline
\end{tabular}
}
\end{table}
%
%
%
\begin{table*}[!ht]
\centering
\caption{WERs in \% of multi-domain evaluation. The best performance is in bold.}
\label{tab_multidomain_full}
\resizebox{0.75\hsize}{!}{%
\begin{tabular}{ccccc}
\hline
Method                                             & $\beta$ & LibriSpeech-test-clean & Common  Voice & TED-LIUM     \\ \hline
ASR baseline1                                      & -    & 2.8                    & 15.3          & 7.0          \\
\multicolumn{1}{l}{+ Stage 1 only: LM re-scoring}  & 0    & 2.5                    & 13.9          & 6.8          \\
\multicolumn{1}{l}{+ Stage 2 only: LLM correction} & 1    & 2.7                    & 13.9          & 6.9          \\
Proposed full pipeline                             & 0.7  & \textbf{2.1}           & \textbf{13.4} & \textbf{6.5} \\ \hline
\end{tabular}
}
\end{table*}
\begin{table*}[!ht]
\centering
\caption{WERs in \% of large-scale ASR evaluation. The best performance is in bold.}
\label{largescale_full}
\resizebox{0.85\hsize}{!}{%
\begin{tabular}{cccc}
\hline
Method                                             & LibriSpeech-test-clean & LibriSpeech-test-other & Multilingual LibriSpeech \\ \hline
MLS  \cite{pratap2020mls}                                              & 1.8                    & 3.5                    & \textbf{5.9}             \\
Wav2vec 2.0 Large \cite{baevski2020wav2vec}                                 & 1.8                    & 3.3                    & -                        \\
Pre-training + Noisy student \cite{zhang2020pushing}                      & 1.4                    & \textbf{2.6}           & -                        \\
Whisper Large-v2 \cite{radford2023robust}                                  & 2.5                    & 4.9                    & 6.2                      \\ \hline
ASR baseline2                                      & 1.6                    & 4.2                    & 6.6                      \\
\multicolumn{1}{l}{+ Stage 1 only: LM re-scoring}  & 1.5                    & 3.9                    & 6.3                      \\
\multicolumn{1}{l}{+ Stage 2 only: LLM correction} & 2.4                    & 4.4                    & 6.2                      \\
Proposed full pipeline                             & \textbf{1.3}           & 3.4                    & 6.0                      \\ \hline
\end{tabular}
}
\end{table*}

\subsection{Finding Hyper-parameters}
Table~\ref{tab_config} shows the WER numbers for different N-best list sizes, the weight $\alpha$ in Equation \ref{equation_rerank}, and the confidence level threshold $\beta$, which are tuning parameters for our correction pipeline. We have tested the beam size $N = [5, 8, 16]$ and tuned the weight $\alpha$ ranged from 1 to 5 with a step size 0.1. The confidence level threshold $\beta$ was tuned with the range from 0 to 1 with a step size 0.05. As we can see from the table, the best configuration is $N=5$, $\alpha=3.0$ and $\beta=0.7$. These numbers will be used and fixed for all other experiments. In the end, only 23\% of total processed speech utterances, whose confidence estimation below the threshold $\beta=0.7$, will be sent to \textit{Stage 2} and handled by a LLM. Majority of utterances (77\%) will be directly outputted from \textit{Stage 1}. This greatly alleviates the total computational cost of the pipeline and reduces overall latency.

\section{Results}
\subsection{Multi-domain Evaluation}
\label{sec_multidomain}
To evaluate how the proposed approach is generalized to different domains, we use an ASR model, called \textit{baseline1}, trained on a mix of three data sets (LibriSpeech, Common Voice and TED-LIUM). WER results are shown in Table~\ref{tab_multidomain_full}. Overall, the proposed correction pipeline significantly improves the ASR performance in term of WER. On LibriSpeech test set, the proposed approach achieves 25\% WER relative improvement over the ASR baseline (2.8 $\to$ 2.1). 

To explore the exact benefit from individual stages, we performed the pipeline with single stages solely, i.e., setting $\beta$ (the confidence level threshold) to be 1 or 0. When $\beta$ is 0, all utterances will pass the first detection stage, i.e. believed to contain no error, and thus no usage of the second LLM correction stage. It only exploits the first stage of traditional language model re-scoring and selects one candidate hypothesis with the highest score as the output, similar to \cite{chen2023large, huang2024multilingual, udagawa2022effect}; When $\beta$ is 1, no utterance can pass the first detection and all of them will be sent to \textit{Stage~2} for LLM corrections. As shown in Table~\ref{tab_multidomain_full}, for \textit{Stage~1} only: the improvement gained by employing the re-scoring with GPT-J is higher or on-par with the use of an in-domain LM, i.e., compared to the results in \cite{park2019specaugment} of similar models and test sets. For \textit{Stage~2} only: the correction with ChatGPT is effective but with a limited effect. This limitation is in line with previous LLM-based correction methods \cite{ma2023can, chen2023hyporadise}.

Combining the results in each individual stage and the full pipeline, we can see that the gains from \textit{Stage~1} and \textit{Stage~2} are complementary to each other. On LibriSpeech, LLM-based correction achieves a large WER relative reduction of 16\% from \textit{Stage 1} to \textit{Stage 2} (2.5 $\to$ 2.1). This clearly shows the benefits of both stages, where the best performance is consistently obtained when performing the full pipeline (\textit{Stage~1} + \textit{Stage~2}).

\subsection{Large-scale ASR Evaluation}
To examine if the proposed approach still works for large-scale ASR, we created a new training data set by merging the LibriSpeech with a large amount of English speech from the Multilingual LibriSpeech corpus. Then we trained a new ASR model, called \textit{baseline2}, with the same size and similar optimizations. 

As shown in Table~\ref{largescale_full}, our large-scale ASR baseline performs well on all the read speech sets, especially on test-clean. Its performance is comparable to the best systems reported in \cite{pratap2020mls, baevski2020wav2vec, zhang2020pushing, radford2023robust}.
When applying single correction stages (similarly to what has been done in Section~\ref{sec_multidomain}), we observed an interesting result. The re-scoring in \textit{Stage~1} still gives consistent improvement but the LLM-based correction in \textit{Stage~2} downgrades the performance on LibriSpeech. We manually reviewed this phenomenon and confirmed that corrections of ChatGPT can produce undesired changes, i.e., the \textit{over-correction} issue. Specifically, its correction would steer sentences more towards written language, instead of the exact verbatim transcriptions of speech. Also, ChatGPT unexpectedly corrected grammar errors but hindered the speech fidelity, raising the WER. This observation reveals the weakness of a solely LLM-based correction.

With the full correction pipeline, the proposed approach is shown to successfully mitigate the over-correction issue. At the end, we can achieve a consistent WER relative improvement of 10$\% \sim$ 20\% cross three test sets in this large-scale ASR benchmark. On LibriSpeech test-clean, our result of  1.3\% WER has made a new state-of-the-art record.

\section{Conclusion}
\label{sec: conclusion}

In this paper, we propose a novel multi-stage approach for LLM-based ASR correction. The proposed approach has been shown to effectively mitigate the over-correction issue of LLMs via an uncertainty estimation stage. Besides, we formulate the ASR correction task as a multi-step rule-based LLM reasoning process, which achieves the SOTA performance in even zero-shot settings. Experimental results show that the proposed approach can provide a 10\% $\sim$ 20\% relative improvement for competitive and multi-domain ASR systems. 

A far-reaching implication of this work is, in the future all interaction and processing of text for ASR will be able to be expressed as rules in  natural language and then handled by LLMs. Yet, the proposed rule-based prompt strongly depends on the reasoning capability of LLMs, especially in zero-shot settings. For the future work, it would be interesting to investigate how to enable the proposed prompt in smaller and less capable LLMs.

\newpage
\bibliographystyle{IEEEtran}
\bibliography{mybib}

\end{document}